\documentclass[10pt,twocolumn,letterpaper]{article}

\usepackage[pagenumbers]{cvpr} % To force page numbers, e.g. for an arXiv version

\usepackage{multirow}

\usepackage{colortbl}
\usepackage{caption}
\newcommand{\method}{\textit{GrndCtrl}}

\usepackage{xcolor}
\definecolor{dimred}{rgb}{0.5, 0.1, 0.1}

\definecolor{dimgreen}{rgb}{0.1, 0.5, 0.1}

\definecolor{dimblue}{rgb}{0.1, 0.1, 0.5}
\makeatletter
\renewcommand{\abstract}{%
   \iftoggle{cvprpagenumbers}{}{%
     \thispagestyle{empty}
   }
   \centerline{\large\bf Abstract}%
   \vspace{6pt}
   \noindent%
   \it\ignorespaces%
}
\makeatother

\definecolor{methodcolor}{HTML}{94b6d2}

\definecolor{cvprblue}{rgb}{0.21,0.49,0.74}
\definecolor{grndctrlgreen}{rgb}{0.3686,0.6118,0.4627}   % #5E9C76
\definecolor{grndctrlauthor}{rgb}{0.2902,0.4941,0.7333}  % #4A7EBB
\definecolor{grndctrlaffil}{rgb}{0.7882,0.3608,0.3294}   % #C95C54
\usepackage[pagebackref,breaklinks,colorlinks,allcolors=cvprblue]{hyperref}

\usepackage{booktabs, tabularx, threeparttable}
\usepackage{multirow}
\usepackage{siunitx}

\def\paperID{21565} % *** Enter the Paper ID here
\def\confName{CVPR}
\def\confYear{2026}

\title{%
  \raisebox{-0.5ex}{\includegraphics[height=2.5ex]{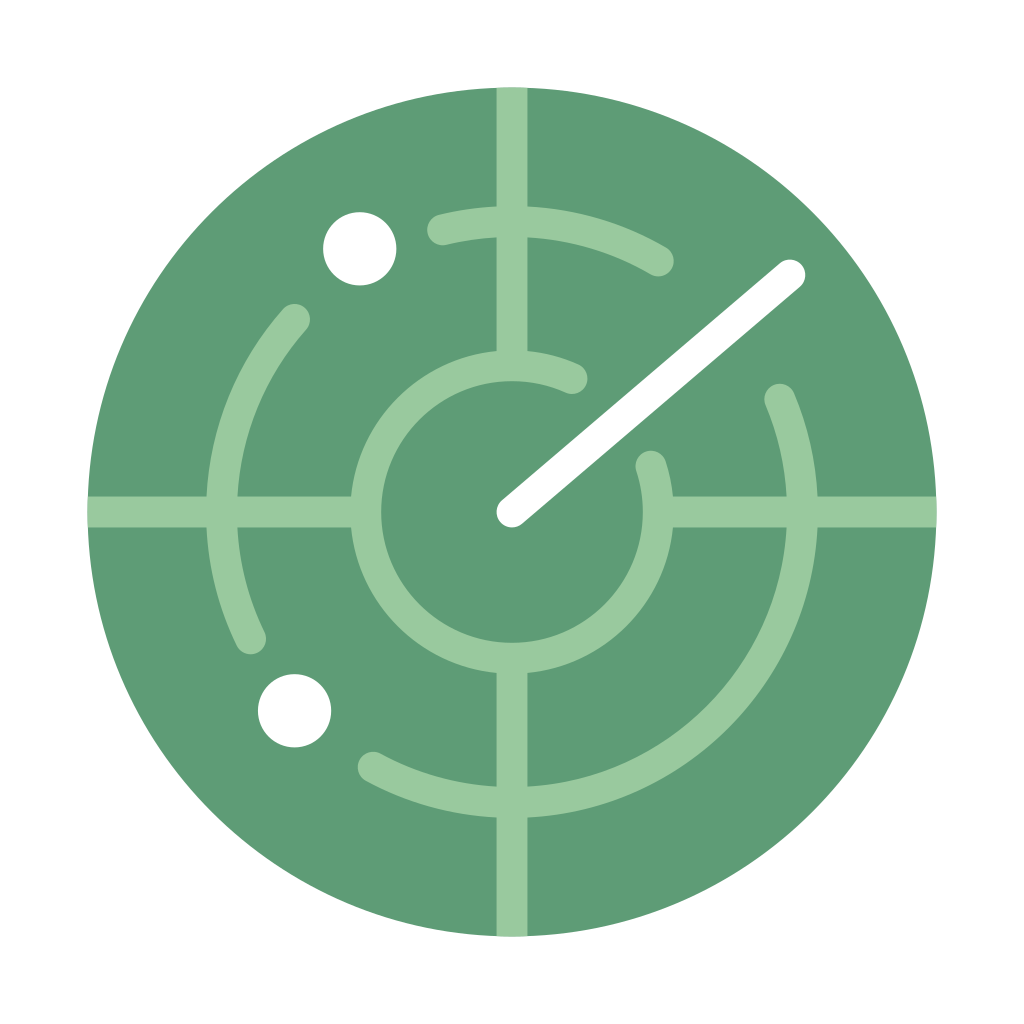}}\,GrndCtrl: Grounding World Models via Self-Supervised Reward Alignment\\[0.4em]
  {\small\href{https://rlwg-grndctrl.github.io/}{\textcolor{grndctrlgreen}{\texttt{https://rlwg-grndctrl.github.io/}}}}%
}

\author{%
\href{https://purenothingness24.github.io/}{\textcolor{grndctrlauthor}{\textbf{Haoyang He}}}$^{1, 2}$\textsuperscript{*}
\and
\href{https://www.jaypatrikar.me/}{\textcolor{grndctrlauthor}{\textbf{Jay Patrikar}}}$^2$
\and
\href{https://dkkim93.github.io/}{\textcolor{grndctrlauthor}{\textbf{Dong\mbox{-}Ki Kim}}}$^2$
\and
\href{https://www.maxosmith.com/}{\textcolor{grndctrlauthor}{\textbf{Max Smith}}}$^2$
\and
\href{https://danmcgann.com/}{\textcolor{grndctrlauthor}{\textbf{Daniel McGann}}}$^2$
\and
\href{https://www.fieldai.com/team}{\textcolor{grndctrlauthor}{\textbf{Ali-akbar Agha-mohammadi}}}$^2$
\and
\href{https://www.fieldai.com/team}{\textcolor{grndctrlauthor}{\textbf{Shayegan Omidshafiei}}}$^2$
\and
\href{https://theairlab.org/team/sebastian/}{\textcolor{grndctrlauthor}{\textbf{Sebastian Scherer}}}$^{1, 2}$
}

\begin{document}

\twocolumn[{%
  % Make \maketitle's internal \twocolumn a no-op so it
  % doesn't start a new page and mess up the layout.
  \renewcommand\twocolumn[1][]{#1}%

  \maketitle

  \iftoggle{cvprfinal}{%
    \vspace{-1cm}
    \begin{center}
      \large
      \noindent$^1$\href{https://www.cmu.edu/}{\textcolor{grndctrlaffil}{\textbf{Carnegie Mellon University}}} \hspace{5em}%
      $^2$\href{https://www.fieldai.com/}{\textcolor{grndctrlaffil}{\textbf{FieldAI}}}
    \end{center}
    \vspace{-0.5cm}
  }{}%

  \begin{center}
    \includegraphics[width=0.8\linewidth]{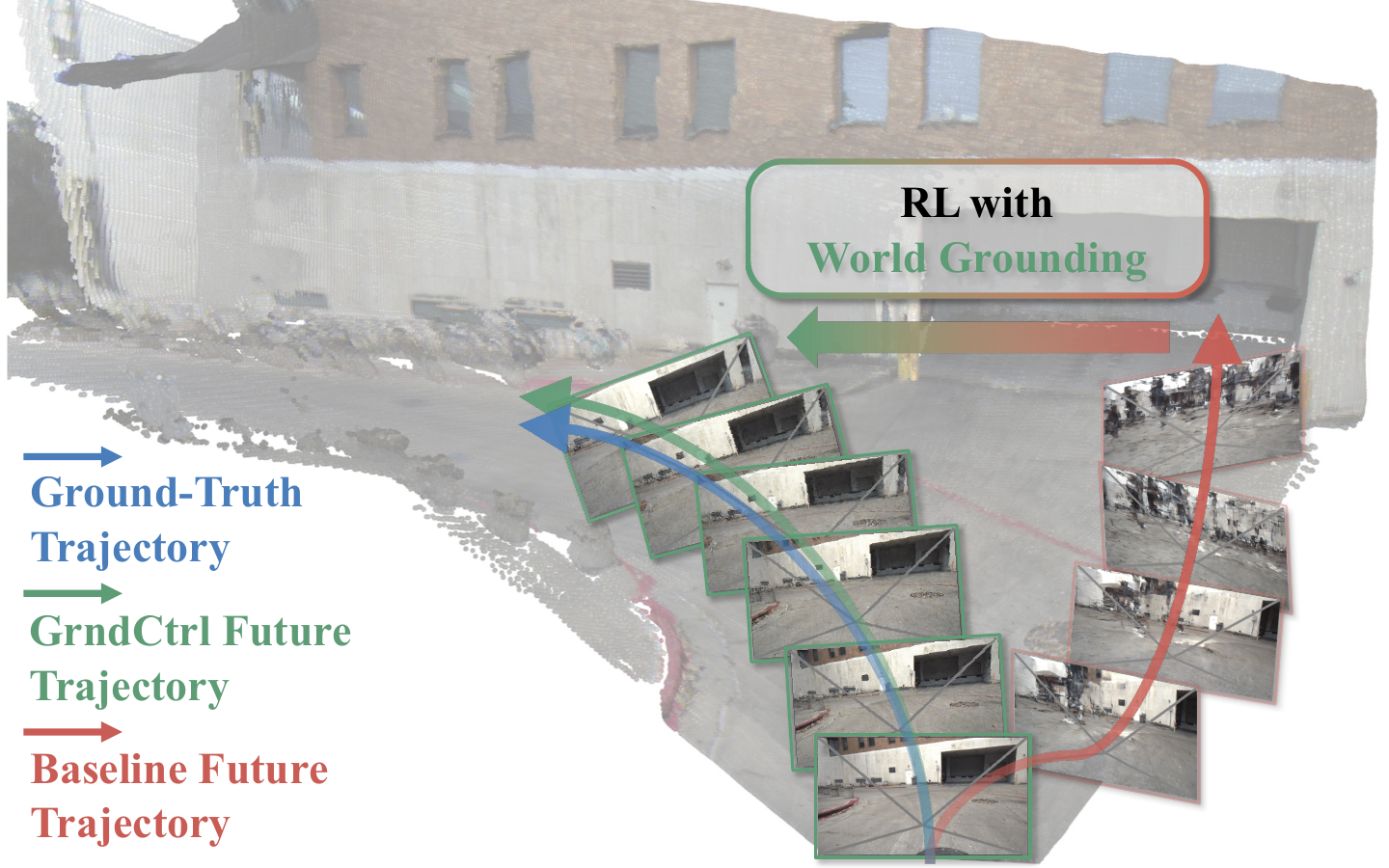}
  \end{center}

  \captionsetup{type=figure}
  \captionof{figure}{%
    \textbf{Reinforcement Learning with World Grounding (RLWG)} addresses geometric inconsistencies in pretrained video world models through self-supervised post-training with verifiable rewards. Instead of reconstruction losses, RLWG grounds models using geometric and perceptual rewards from frozen evaluators. \method{} instantiates RLWG using Group Relative Policy Optimization (GRPO), enabling physically consistent rollouts essential for reliable world generation.}
  \label{fig:teaser}
  \vspace{0.25cm}
}]

\makeatletter
\begingroup
\def\@thefnmark{*}\@footnotetext{Work done while intern at FieldAI.}
\endgroup
\makeatother

\begin{abstract}
Recent advances in video world modeling have enabled large-scale generative models to simulate embodied environments with high visual fidelity, providing strong priors for prediction, planning, and control. Yet, despite their realism, these models often lack geometric grounding, limiting their use in navigation tasks that require spatial coherence and stability. We introduce \textbf{Reinforcement Learning with World Grounding (RLWG)}, a self-supervised post-training framework that aligns pretrained world models with a physically verifiable structure through geometric and perceptual rewards. Analogous to reinforcement learning from verifiable feedback (RLVR) in language models, RLWG can use multiple rewards that measure pose cycle-consistency, depth reprojection, and temporal coherence. We instantiate this framework with \textbf{GrndCtrl}, a reward-aligned adaptation method based on Group Relative Policy Optimization (GRPO), yielding world models that maintain stable trajectories, consistent geometry, and reliable rollouts for embodied navigation. Like post-training alignment in large language models, \textbf{GrndCtrl} leverages verifiable rewards to bridge generative pretraining and grounded behavior, achieving superior spatial coherence and navigation stability over supervised fine-tuning in outdoor environments.
\vspace{-0.7cm}
% Website: \hyperlink{grndctrl.github.io}{grndctrl.github.io}
\end{abstract}
\section{Introduction}
\label{sec:intro}
Large-scale video world models have emerged as powerful priors for modeling perception and control for embodied agents~\cite{nvidia2025cosmosworldfoundationmodel, bar2025navigationworldmodels, assran2025vjepa2, jin2025posepilotsteeringcamerapose, ren2025gen3c}. By learning to predict future observations from past frames and actions, these models approximate the transition dynamics of the physical world, enabling simulation, planning, and policy evaluation. Operating in the pixel domain aligns them with real-world sensors and exploits the vast implicit supervision available in video, allowing unified modeling across domains such as manipulation, driving, and navigation. Yet despite their impressive generative fidelity, these models are often incentivized to capture the appearance of motion more than its structure. Their rollouts remain visually plausible but geometrically and temporally inconsistent: poses drift, depths wobble, and trajectories lose alignment over time. Even subtle deviations in inferred geometry accumulate into compounding spatial errors corrupting metric structure. These instabilities limit the use of current models for closed-loop tasks such as localization, mapping, and planning, where physically consistent representation is essential.

We define \textbf{world model grounding} as aligning learned dynamics with physically verifiable spatial and temporal invariants, so that rollouts honor geometry and time in addition to reproducing surface appearance. Grounding shifts the objective of world modeling from visual plausibility to structural consistency, ensuring that the model’s internal dynamics respect the constraints of real motion and scene structure. To this end, we introduce \textbf{Reinforcement Learning with World Grounding (RLWG)}, a self-supervised post-training framework that refines pretrained world models using verifiable geometric and perceptual rewards derived from model rollouts. RLWG extends the principle of Reinforcement Learning with Verifiable Rewards (RLVR) from language models~\cite{lambert2025tulu3pushingfrontiers} to the embodied domain, replacing text-based logical verification with geometric and temporal verification. In RLWG, a pretrained world model is treated as a policy that generates multiple candidate rollouts from the same context; each rollout is automatically scored using verifiable grounding rewards that quantify spatial and temporal coherence, such as pose cycle-consistency, depth reprojection agreement and action adherence. Unlike reconstruction losses that only penalize pixel error, these rewards measure physical correctness of the rollouts.

To optimize these verifiable rewards efficiently, we adopt Group Relative Policy Optimization (GRPO)~\cite{shao2024deepseekmathpushinglimitsmathematical} as our training mechanism, yielding our algorithm, \method. For each context (and actions when available), the model generates a group of rollouts that are ranked by their grounding rewards; relative advantages are computed within the group, and the latent transition operator is updated using a clipped policy gradient objective regularized toward the pretrained model. This formulation preserves visual quality while progressively aligning the model’s dynamics with measurable structure in the real world. The process requires no human annotations or external simulators, operating entirely through self-supervised reinforcement grounded in the model’s own predictions. Conceptually, \method \  extends the success of GRPO-based alignment in generative modeling to the geometric domain, grounding visual world models in verifiable 3D and temporal coherence.

This paradigm reframes the role of post-training in world modeling. Rather than optimizing for perceptual fidelity or next-frame likelihood, RLWG drives the model toward internal representations that are self-consistent and physically grounded. It establishes a structural analogue to the self-alignment processes that have improved reasoning in large language models: where RLVR grounds language in logic, RLWG grounds world models in geometry. The resulting models are self-grounded, spatially coherent, and dynamically stable—capable not only of rendering the world vividly, but of representing it in actionable, physically consistent form. Through this lens, we move beyond visually coherent generation toward structurally consistent simulation, bridging the gap between generative video modeling and physical world understanding, and opening a path toward world models that can both imagine and inhabit the real world.

The main contributions of this work are:
\begin{enumerate}
    \item We introduce \textbf{Reinforcement Learning with World Grounding (RLWG)}, a self-supervised grounding framework using verifiable geometric and temporal rewards from frozen evaluators without labels or simulators.
    \item We construct \method, a method that extends GRPO to the RLWG regime by  multi-reward alignment over stochastic rollouts optimizing Translation, Rotation, Depth Temporal Reprojection Inlier ratio, and perceptual quality with pretrained frozen evaluators.
    \item We provide a comprehensive evaluation of \method \   across multiple datasets showing reduced pose error means and variances, with strong gains under counterfactual rollouts and generalization to unseen inputs.
\end{enumerate}

% \clearpage

\section{Related Work}

\subsection{Controlling World Models}
Recent progress in large-scale video foundation models has transformed video prediction into controllable world simulation. Models such as Cosmos-Predict~\cite{nvidia2025cosmosworldfoundationmodel}, and V-JEPA~\cite{assran2025vjepa2} unify multi-modal conditioning for long-horizon prediction and control. These models achieve impressive simulation fidelity but still exhibit spatial drift, geometric misalignment, and temporal incoherence over extended rollouts, revealing limitations in geometric grounding. Architectural innovations like flow matching, conditional diffusion transformers, and masked latent prediction have improved realism but not physical consistency.  

Controlling these models can be categorized into \textit{action-conditioned} and \textit{camera-conditioned} paradigms. The first paradigm trains models to predict futures from discrete actions, and can be further defined by the nature of the action and observation frame. Some methods~\cite{zhou2025dinowmworldmodelspretrained, zhu2025unifiedworldmodelscoupling} assume a static camera observing an embodiment performing actions, while others~\cite{hafner2024masteringdiversedomainsworld, valevski2025diffusionmodelsrealtimegame, bar2025navigationworldmodels, bai2025wholebodyconditionedegocentricvideo} assume a fixed camera on the embodiment and train models to predict ego-centric views. Jointly, they aim to  enable model-predictive control, but face challenges with physical realism. The second paradigm explicitly decouples viewpoint from embodiment, allowing the model to generate future observations from arbitrary camera poses. Early works~\cite{wang2024motionctrlunifiedflexiblemotion, he2025cameractrlenablingcameracontrol} injected pose embeddings into diffusion models to achieve this control, but lacked explicit geometric alignment. Subsequent methods~\cite{jin2025posepilotsteeringcamerapose,ren2025gen3c} improved consistency via self-supervised warping, 3D-informed point cloud conditioning, achieving more precise viewpoint control. Recent work~\cite{guo2025ctrlworldcontrollablegenerativeworld} bridges the two paradigms by jointly training on multiple viewpoints, and improved spatial-temporal consistency with pose-conditioned memory retrieval. Despite these advances, most approaches rely on supervised learning or one-step consistency objectives and remain open-loop, with no mechanism to evaluate or optimize physical correctness. 
\subsection{Reward Learning for Post-Training}
Reinforcement-based post-training has become central to aligning large generative models. In language systems, Reinforcement Learning from Human Feedback (RLHF)~\cite{ouyang2022traininglanguagemodelsfollow} and Reinforcement Learning with Verifiable Rewards (RLVR)~\cite{lambert2025tulu3pushingfrontiers} replace imitation with objective-driven alignment, while Group Relative Policy Optimization (GRPO)~\cite{shao2024deepseekmathpushinglimitsmathematical} stabilizes learning via stochastic rollouts comparative updates. Extensions to vision, such as DanceGRPO~\cite{xue2025dancegrpounleashinggrpovisual}, demonstrate that rewards on visual quality can fine-tune video diffusion models effectively.

Building on this foundation, RLWG adapts RLVR to world modeling, optimizing pretrained world models using physically verifiable rewards including cycle-consistency, depth reprojection, and trajectory stability. \method \ instantiates RLWG as a multi-objective GRPO over grounded rewards. This process enforces geometric coherence without human supervision, significantly reducing scene drift. Parallel advances in 3D perception emphasize similar constraints: VGGT~\cite{wang2025vggt} and MapAnything~\cite{keetha2025mapanythinguniversalfeedforwardmetric} predict depth and camera pose for consistent scene reconstruction, while SpaTracker~\cite{xiao2025spatialtrackerv23dpointtracking} integrates rigidity priors for robust 3D tracking. Together, these efforts point toward reward-informed geometry as a unifying principle, where physical correctness acts as an alignment signal bridging generative modeling, simulation, and control.

\begin{figure*}
    \centering
    \includegraphics[width=0.95\textwidth]{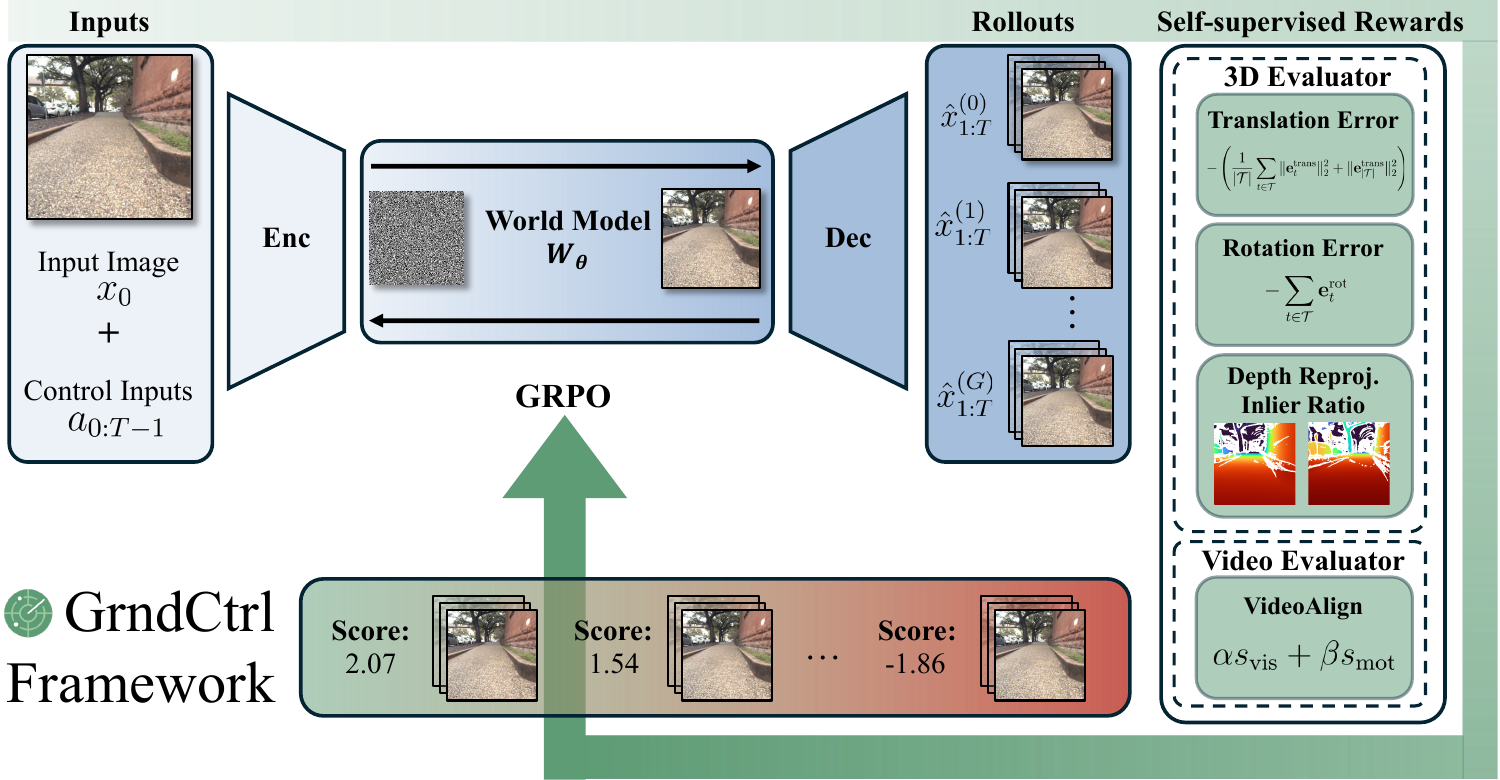}
    \caption{\method{} framework architecture. Given conditioning context $c = (x_0, a_{0:T-1})$, the world model generates multiple stochastic rollouts $\{\hat{x}^{(i)}_{1:T}\}_{i=1}^{G}$. Frozen evaluators compute verifiable rewards for each rollout. Relative advantages are calculated within each group, and GRPO updates model parameters using a clipped policy gradient objective regularized toward the pretrained model, favoring physically consistent rollouts.}
    \label{fig:framework}
\end{figure*}

\section{GrndCtrl}

\subsection{Problem Definition} \label{sec:method:problem-definition}

We consider a pretrained \emph{video world model} $W_\theta$, a policy parameterized by $\theta$, that predicts future observations conditioned on a visual history and optionally actions. 
Let $x_0$ denote the observed frame and $a_{0:T-1} = (a_0, \ldots, a_{T-1})$ the associated control inputs.
The model samples a rollout $\hat{x}_{1:T} \sim W_\theta(\cdot \mid x_0, a_{0:T-1})$,
where $a_t = (R_t, \mathbf{t}_t)$ controls translation $\mathbf{t}_t \in \mathbb{R}^3$ and rotation $ R_t \in \mathrm{SO}(3)$. Our goal is to \emph{post-train} $W_\theta$ using self-supervised reinforcement learning to improve the \emph{spatial coherence} and \emph{embodied reliability} of its rollouts.

To obtain \emph{verifiable} feedback without supervision, we use a frozen \emph{feed-forward 3D evaluator} $\mathcal{E}$ that provides relative pose estimates $(\Delta R_{1:T}, \Delta \mathbf{t}_{1:T})$ and per-frame depth maps $D_t$,
where $\Delta R_t \in \mathrm{SO}(3)$ and $\Delta \mathbf{t}_t \in \mathbb{R}^3$. Additionally, we obtain feedback on the visual and motion quality of the overall video using a frozen \emph{feed-forward video evaluator} $\mathcal{V}$ that provides overall visual quality scoring.

Our objective is to optimize $\theta$ such that sampled rollouts maximize a set of verifiable rewards $U(\hat{x}_{1:T})$ comprising translation ($r_{\text{trans}}$), rotation ($r_{\text{rot}}$), depth temporal reprojection ($r_{\text{dtr}}$), and video quality ($r_{\text{v}}$).
These rewards are constructed from the 3D evaluator $\mathcal{E}$ and video evaluator $\mathcal{V}$, and are defined in detail in Section~\ref{sec:method:verifiable-rewards}.

\subsection{Verifiable Self-Supervised Rewards} \label{sec:method:verifiable-rewards}

Let $\mathcal{T}$ denote the set of evaluated timesteps.  
Each reward term measures a distinct aspect of spatial and temporal consistency.

\paragraph{(1) Translation Reward.}
We compute the Euclidean deviation in translation:
\begin{equation}
    \mathbf{e}^{\text{trans}}_t = \Delta \mathbf{t}_t - \mathbf{t}_t.
\end{equation}
and define the sum of mean squared trajectory error and squared final error as translation reward: 
\begin{equation}
    r_{\text{trans}} = -\left( \frac{1}{|\mathcal{T}|}\sum_{t \in \mathcal{T}}\|\mathbf{e}^{\text{trans}}_t\|^2_2 + \|\mathbf{e}^{\text{trans}}_{|\mathcal{T}|}\|^2_2 \right).
\end{equation}
When metric scale is ambiguous, a normalization factor is applied from the evaluator's scale estimate.

\paragraph{(2) Rotation Reward.}
We compute the minimum angular deviation between predicted and evaluator rotations:
\begin{equation}
    \mathbf{e}^{\text{rot}}_t = \arccos \left( \frac{\operatorname{tr}\left(\Delta R_t \cdot {R_t}^{T}\right) - 1}{2} \right),
\end{equation}
and define the axis-angle cumulative error as rotation reward:
\begin{equation}
    r_{\text{rot}} = -\sum_{t\in\mathcal{T}}\mathbf{e}^{\text{rot}}_t.
\end{equation}

\paragraph{(3) Depth Temporal Reprojection Reward.}
We adopt the \textbf{depth inlier ratio} from MapAnything~\cite{keetha2025mapanythinguniversalfeedforwardmetric} evaluations as a verifiable geometric reward for depth temporal reprojection.
For each pixel $p\!\in\!\Omega$, define the reprojected correspondence and expected depth via the evaluator geometry
\begin{equation}
(\hat{p},\, d_{t\to t+1}^{\exp}(p)) \;=\; \Phi\!\big(p;\, D_t,\, \Delta R_t,\, \Delta \mathbf{t}_t,\, K_t,K_{t+1}\big),
\end{equation}
where $K_t,K_{t+1}$ are the intrinsics of frames $t$ and $t{+}1$, and $d_{t\to t+1}^{\exp}(p)$ is the depth at $t{+}1$ obtained by back-projecting $p$ with $D_t(p)$, transforming by $(\Delta R_t,\Delta\mathbf{t}_t)$, and re-projecting to $\hat p$.
The per-pair depth inlier ratio (threshold $\gamma{=}0.0103$) is
\begin{equation}
\mathrm{DTRI}_t^{(\gamma)} \;=\; \frac{1}{|\Omega|}\sum_{p\in\Omega}
\mathbf{1}\!\left[
\frac{\big|\,D_{t+1}(\hat{p}) - d_{t\to t+1}^{\exp}(p)\,\big|}{d_{t\to t+1}^{\exp}(p)} \;<\; \gamma
\right].
\end{equation}
We use the average inlier ratio as the reward:
\begin{equation}
r_{\text{dtr}} \;=\; \frac{1}{|\mathcal{T}|}\sum_{t\in\mathcal{T}} \mathrm{DTRI}_t^{(\gamma)}.
\end{equation}

\paragraph{(4) Video Quality Reward.}
We use the frozen VideoAlign~\cite{liu2025improving} as evaluator $\mathcal{V}$, which returns three sequence-level scores in $[0,1]$—visual quality $s_{\text{vis}}$, motion quality $s_{\text{mot}}$, and text alignment $s_{\text{txt}}$—for a rollout $\hat{x}_{1:T}$ (optionally conditioned on a prompt $y$):
\begin{equation}
(s_{\text{vis}},\, s_{\text{mot}},\, s_{\text{txt}}) \;=\; \mathcal{V}(\hat{x}_{1:T};\, y).
\end{equation}
Our visual reward uses only visual and motion quality as a convex combination,
\begin{equation}
r_{\text{v}} \;=\; \alpha\, s_{\text{vis}} + \beta\, s_{\text{mot}}, 
\qquad \alpha,\beta \ge 0,\;\alpha+\beta=1,
\end{equation}
with $\alpha=\beta=\tfrac{1}{2}$ by default.

\subsection{GRPO for RLWG Post-Training} \label{sec:method:GRPO}

For each \emph{conditioning context} $c = (x_0, a_{0:T-1})$, a group of $G$ candidate rollouts $\{\hat{x}^{(i)}_{1:T}\}_{i=1}^{G}$ is sampled from $W_\theta$. 
Each rollout is evaluated by every reward from the verifiable reward set $U(\hat{x}_{1:T})$, obtaining $\{r^{(i)}_{\text{trans}}, r^{(i)}_{\text{rot}}, r^{(i)}_{\text{dtr}}, r^{(i)}_{\text{v}} \} = U(\hat{x}^{(i)}_{1:T})$. We compute the normalized reward for every verifiable reward $\tilde{r}^{(i)}$, and obtain a multi-objective normalized group advantage $A_i$:

\begin{equation}
    \tilde{r}^{(i)} = \tfrac{r^{(i)} - \text{mean}(r^{(1)}, \ldots, r^{(G)})}{\text{std}(r^{(1)}, \ldots, r^{(G)})}, \quad 
    A_i = \tfrac{\tilde{r}^{(i)} - \text{mean}(\tilde{r}^{(1)}, \ldots, \tilde{r}^{(G)})}{\text{std}(\tilde{r}^{(1)}, \ldots, \tilde{r}^{(G)})},
\end{equation}

The GRPO objective optimizes $\theta$ using the clipped surrogate:
\begin{equation}
    J(\theta) =
    \mathbb{E}\Big[
    \frac{1}{G}\sum_{i=1}^{G}\frac{1}{T}\sum_{t=1}^{T}
    \min\big( \rho_{t,i} A_i,\; \text{clip}(\rho_{t,i}, 1-\epsilon, 1+\epsilon) A_i \big)
    \Big],
\end{equation}
where $\epsilon$ is the clip ratio, $\rho_{t,i} $ is the per-step likelihood ratio under the current and reference policies.
This formulation stabilizes policy optimization for continuous video generation tasks. An illustration of the detailed \method{} framework is shown in Figure~\ref{fig:framework}. To form each group for GRPO in practice, we generate multiple stochastic rollouts.

% \paragraph{Stochastic rollouts}
Groups are instantiated by sampling multiple candidate rollouts from the diffusion generator under controlled stochasticity.\label{sec:method:stochastic-rollouts} Diffusion sampling can be formulated as a reverse-time \emph{stochastic differential equation} (SDE) or, equivalently in time-marginals, as a \emph{probability–flow ordinary differential equation} (ODE)~\cite{song2021scorebasedgenerativemodelingstochastic, albergo2025stochasticinterpolantsunifyingframework}. Let $x_t\in\mathbb{R}^d$ denote the latent state at time $t \in [0,1]$ (noisiest at $t{=}1$, data manifold at $t{\to}0$). A standard variance–preserving forward SDE is
\begin{equation}
\mathrm{d}x_t  = -\tfrac{1}{2}\beta(t)x_t\mathrm{d}t  + \sqrt{\beta(t)}\mathrm{d}w_t,
\label{eq:vp-forward}
\end{equation}
with $w_t$ as a standard Brownian motion and $\beta(t){>}0$ as the noise rate. The corresponding \emph{reverse} SDE is
\begin{equation}
\mathrm{d}x_t  = \big(-\tfrac{1}{2}\beta(t)x_t  - \beta(t)\nabla_x\log p_t(x_t)\big)\mathrm{d}t  + \sqrt{\beta(t)}\mathrm{d}\bar w_t,
\label{eq:vp-reverse}
\end{equation}
which is stochastic due to the $\mathrm{d}\bar w_t$ term and uses the learned score $\nabla_x\log p_t$, where $p_t$ is the forward SDE solution distribution of $x_t$. The \emph{probability–flow ODE}, which shares the same ${p_t}$ as \eqref{eq:vp-reverse} but is deterministic, is
\begin{equation}
\frac{\mathrm{d}x_t}{\mathrm{d}t}  = -\tfrac{1}{2}\beta(t)x_t  - \tfrac{1}{2}\beta(t)\nabla_x\log p_t(x_t).
\label{eq:pf-ode}
\end{equation}
In practice, sampler stochasticity is controlled by $\eta \in [0,1]$ (ODE limit at $\eta{=}0$; SDE-style sampling with per-step Gaussian perturbations at $\eta{=}1$). For each context $c$, we draw $G$ rollouts $\{\hat{x}_{1:T}^{(i)}\}_{i=1}^{G}$ with identical conditioning and independent noise governed by $\eta$, providing the within-group diversity required by GRPO.

\section{Experimental Setup}
\begin{figure*}[h!]
  \centering
  % (a) Scene drift
  \begin{subfigure}[t]{\textwidth}
    \centering
    \includegraphics[width=\textwidth,page=1]{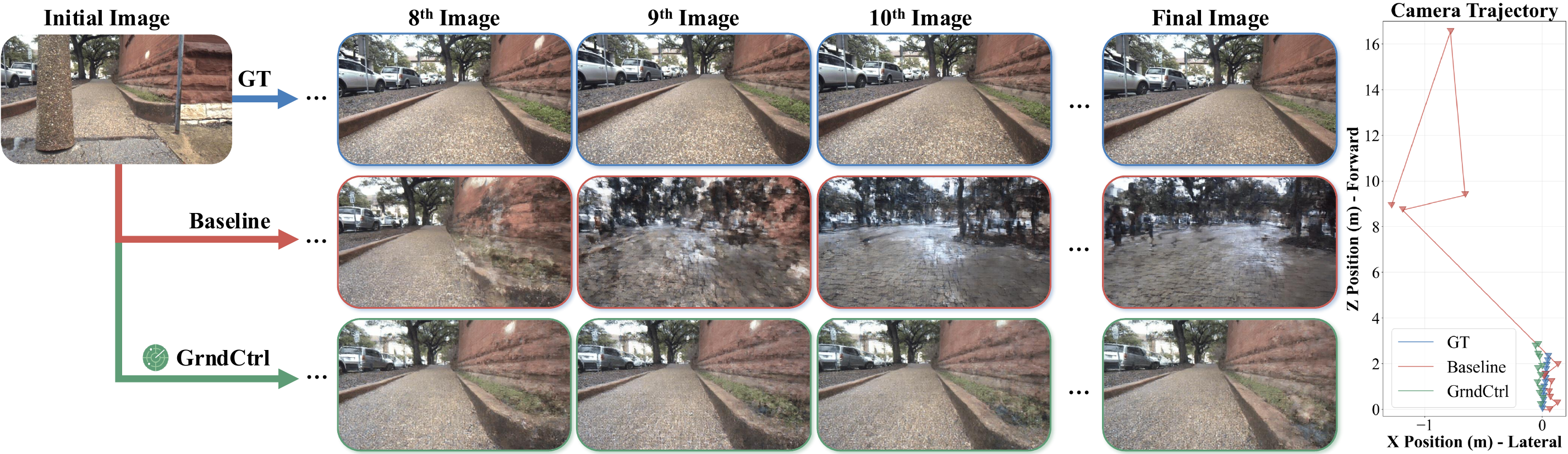}
    \subcaption{}
    \label{subfig:qualitative_drift}
  \end{subfigure}

  % \vspace{0.75em}

  % (b) Trajectory following
  \begin{subfigure}[t]{\textwidth}
    \centering
    \includegraphics[width=\textwidth,page=2]{figures/grndctrl_figure3.pdf}
    \subcaption{}
    \label{subfig:qualitative_direction}
  \end{subfigure}

  \caption{Qualitative results: (a) \method{} mitigates scene drift on counterfactual rollouts, maintaining spatial coherence where baseline diverges. (b) \method{} successfully follows directionally inverted actions, generating geometrically consistent rollouts where baseline fails.}
  \label{fig:qualitative_combined}
\end{figure*}
\subsection{Datasets}
We train and evaluate our methods on three datasets spanning diverse embodiments and scenarios. \textbf{CODa}~\cite{zhang2023towards} is a campus navigation dataset collected on wheeled robots with pseudo-ground truth poses. \textbf{SCAND}~\cite{karnan2022sociallycompliantnavigationdataset} is a social navigation dataset also collected on campus, featuring embodiments of both a wheeled robot and a quadruped. \textbf{CityWalk}~\cite{liu2025citywalker} is an egocentric urban navigation dataset of a person walking in crowded city streets, collected from in-the-wild YouTube city walking videos and reprocessed with MapAnything~\cite{keetha2025mapanythinguniversalfeedforwardmetric} to obtain pose estimates. We subsample each dataset to 2k non-overlapping 13-frame sequences of pose-action pairs for training, ensuring temporal diversity and avoiding data leakage.

\subsection{Baseline Model}
We obtain the baseline pretrained world model $W_\theta$ via supervised fine-tuning (SFT) on Cosmos-Predict2-2B-Video2World~\cite{nvidia2025cosmosworldfoundationmodel}, a diffusion model with a latent VAE backbone and temporal attention. We use a modified version of its action-conditioned video predictor post-training pipeline, adapting the action space from 7 to 6 degrees of freedom to match our navigation setting: $(x, y, z, \mathrm{roll}, \mathrm{pitch}, \mathrm{yaw})$ as 6×12 action embeddings. We initialize the action embeddings randomly while fine-tuning the entire DiT backbone for 20k steps with an effective batch size of 64. Training follows the EDM framework~\cite{karras2022elucidatingdesignspacediffusionbased, karras2024analyzingimprovingtrainingdynamics}, using the weighted expectation of denoising score matching loss over noise levels. We perform full SFT over all DiT backbone parameters keeping the visual encoder and decoder frozen, as our experiments showed that only full SFT or full-size LoRA yielded meaningful changes in motion dynamics. Additional implementation details are provided in Appendix.

\subsection{GRPO Post-Training}
Post-training applies GRPO with self-supervised verifiable rewards as described in Sec.~\ref{sec:method:GRPO}. We use MapAnything~\cite{keetha2025mapanythinguniversalfeedforwardmetric} as $\mathcal{E}$ to obtain rewards $r_{\text{trans}}$, $r_{\text{rot}}$, and $r_{\text{dtr}}$. We use VideoAlign~\cite{liu2025improving} as $\mathcal{V}$ to obtain the reward $r_{\text{v}}$. We perform an ablation study of combinatorial rewards as defined in Sec.~\ref{sec:method:verifiable-rewards} to evaluate multi-objective GRPO. For each reward configuration, we train for 100 steps with an effective batch size of 8, generating $G=8$ stochastic rollouts $\{\hat{x}^{(i)}_{1:T}\}_{i=1}^{G}$ per conditioning context $c = (x_0, a_{0:T-1})$. To obtain diverse rollouts, we use the same initial noise for reverse-SDE diffusion with the same starting frame $x_0$ and action trajectory $a_{0:T-1}$, but inject stochastic Brownian noise at each diffusion step as described in Sec.~\ref{sec:method:stochastic-rollouts}. We compute per-step likelihood ratios only over the first 60\% of diffusion timesteps to focus training on the most relevant denoising steps to improve training stability. Additional training details are provided in Appendix.

\subsection{Evaluation Regimes}
We evaluate on three regimes that progressively test generalization capabilities:

\begin{itemize}
    \item \textbf{Seen}: Start frames and scene domains seen during SFT with matching action distributions. This regime tests in-distribution performance.
    \item \textbf{Counterfactual}: Scenes seen during SFT but with counterfactual actions (e.g., mirrored or directionally inverted action sequences). This regime tests the model's ability to extrapolate geometric structure to novel motion patterns.
    \item \textbf{Unseen}: Both scenes and actions novel at test time, with scenes from different domains and action sequences with different motion characteristics. This regime tests full generalization to unseen scenarios.
\end{itemize}

\subsection{Metrics}
We report four metrics averaged across an evaluation set of 200 non-overlapping sequences for each regime: Translation error ($\text{T} = -r_{\text{trans}}$) in meters, Rotation error ($\text{R} = -r_{\text{rot}}$) in radians, Video Quality ($\text{V} = r_{\text{v}}$) combining visual and motion quality scores, and Depth Temporal Reprojection Inlier Ratio ($\text{DTRI} = r_{\text{dtr}}$) as a percentage. Lower values indicate better performance for T and R, while higher values are better for V and DTRI.

\section{Analysis and Discussion}
\label{sec:results}

\begin{table*}[h!]
\centering
\begin{threeparttable}
\setlength{\tabcolsep}{4pt}
\renewcommand{\arraystretch}{1.15}

% Column pattern per split: T (3.1), R (1.2), V (1.2), DTRI (2.1)
\begin{tabular}{l
  S[table-format=3.1] S[table-format=1.2] S[table-format=1.2] S[table-format=2.1]
  S[table-format=3.1] S[table-format=1.2] S[table-format=1.2] S[table-format=2.1]
  S[table-format=3.1] S[table-format=1.2] S[table-format=1.2] S[table-format=2.1]}
\toprule
& \multicolumn{4}{c}{\textbf{Seen}} & \multicolumn{4}{c}{\textbf{Counterfactual}} & \multicolumn{4}{c}{\textbf{Unseen}} \\
\cmidrule(lr){2-5}\cmidrule(lr){6-9}\cmidrule(lr){10-13}
\textbf{Method}
& \multicolumn{1}{c}{$T\downarrow$}
& \multicolumn{1}{c}{$R\downarrow$}
& \multicolumn{1}{c}{$V\uparrow$}
& \multicolumn{1}{c}{DTRI$\uparrow$}
& \multicolumn{1}{c}{$T\downarrow$}
& \multicolumn{1}{c}{$R\downarrow$}
& \multicolumn{1}{c}{$V\uparrow$}
& \multicolumn{1}{c}{DTRI$\uparrow$}
& \multicolumn{1}{c}{$T\downarrow$}
& \multicolumn{1}{c}{$R\downarrow$}
& \multicolumn{1}{c}{$V\uparrow$}
& \multicolumn{1}{c}{DTRI$\uparrow$} \\
\midrule
\multicolumn{13}{c}{\textbf{CODa}~\cite{zhang2023towards}} \\
\addlinespace[0.25em]
Baseline \cite{nvidia2025cosmosworldfoundationmodel}
& 57.8 & 1.77 & 7.40 & 38.9
& 71.5 & 1.55 & 7.41 & 39.1
& 56.9 & 1.71 & 7.40 & 38.3 \\
{+}T{+}R
& 46.4 & 1.44 & 7.32 & 38.4
& 50.5 & 1.53 & 7.34 & 38.7
& 54.3 & 1.75 & 7.36 & 39.3 \\
{+}T{+}R{+}DTRI
& 65.7 & 1.74 & 7.43 & 37.0
& 57.7 & 1.86 & 7.42 & 36.8
& 42.6 & 1.74 & 7.40 & 37.1 \\
\rowcolor{methodcolor!30}
{+}T{+}R{+}DTRI{+}V
& 39.9 & 1.27 & 7.35 & 37.5
& 40.7 & 1.42 & 7.34 & 37.4
& 31.0 & 1.53 & 7.37 & 38.0 \\
\addlinespace[0.3em]
\midrule
\multicolumn{13}{c}{\textbf{SCAND}~\cite{karnan2022sociallycompliantnavigationdataset}} \\
\addlinespace[0.25em]
Baseline \cite{nvidia2025cosmosworldfoundationmodel}
& 186.3 & 3.76 & 7.16 & 23.6
& 315.9 & 4.24 & 7.13 & 21.4
& 117.0 & 4.02 & 6.99 & 18.4 \\
{+}T{+}R
& 158.2 & 3.61 & 7.19 & 23.7
& 251.2 & 4.34 & 7.18 & 21.7
& 131.1 & 3.95 & 7.04 & 19.1 \\
{+}T{+}R{+}DTRI
& 157.9 & 3.65 & 7.10 & 22.1
& 288.6 & 4.45 & 7.17 & 20.1
& 118.6 & 4.07 & 7.03 & 17.9 \\
\rowcolor{methodcolor!30}
{+}T{+}R{+}DTRI{+}V
& 133.4 & 3.30 & 7.11 & 24.5
& 220.1 & 4.23 & 7.08 & 22.8
& 123.4 & 3.62 & 6.98 & 19.4 \\
\addlinespace[0.3em]
\midrule
\multicolumn{13}{c}{\textbf{CityWalk}~\cite{liu2025citywalker}} \\
\addlinespace[0.25em]
Baseline \cite{nvidia2025cosmosworldfoundationmodel}
& 11.7 & 3.13 & 7.96 & 46.9
& 13.1 & 3.27 & 7.94 & 47.4
& 20.8 & 4.47 & 7.90 & 44.5 \\
{+}T{+}R
& 8.9 & 3.31 & 7.90 & 44.9
& 4.8 & 4.42 & 7.91 & 45.6
& 10.2 & 3.47 & 7.87 & 42.8 \\
{+}T{+}R{+}DTRI
& 8.4 & 3.36 & 7.84 & 43.5
& 4.7 & 4.40 & 7.83 & 44.1
& 10.9 & 3.68 & 7.79 & 41.4 \\
\rowcolor{methodcolor!30}
{+}T{+}R{+}DTRI{+}V
& 8.8 & 3.37 & 7.84 & 42.6
& 4.7 & 4.37 & 7.85 & 43.3
& 9.9 & 3.74 & 7.80 & 40.8 \\
\bottomrule
\end{tabular}

\vspace{-7pt}
\captionsetup{font=footnotesize}
\caption{Quantitative evaluation across three datasets (\textbf{CODa}, \textbf{SCAND}, \textbf{CityWalk}) and three regimes: \textbf{Seen}, \textbf{Counterfactual}, and \textbf{Unseen}. We compare baseline against progressive reward combinations (T+R, T+R+DTRI, T+R+DTRI+V). \method{} achieves substantial improvements. Metrics: T (Translation Error, m), R (Rotation Error, rad), V (Video Quality), DTRI (Depth Temporal Reprojection Inliers).}
\label{tab:results}
\end{threeparttable}
\end{table*}

\subsection{Impact of \method{} on World Model Failures}
% \paragraph{O1: \method{} mitigates scene drift and improves rollout consistency.}
% Figure~\ref{fig:qualitative_combined} showcases qualitative comparisons of \method{} rollouts $\hat{x}_{1:T}$ against the baseline. \method{} significantly mitigates the scene drift failures frequently observed when queried with mirrored input action sequences $a_{0:T-1}$, and improves rollout consistency.

\paragraph{O1: Baselines show poor counterfactual performance but good generalization within familiar motion manifolds.}
Table~\ref{tab:results} reports quantitative results on CODa, SCAND, and CityWalk datasets across three evaluation regimes. Across all datasets, the baseline performs well in Seen but degrades significantly in Counterfactual, indicating limited transfer to out-of-distribution action sequences. On CODa, baseline translation error increases by $24\%$ from Seen to Counterfactual, while on SCAND the degradation is more severe, increasing by $70\%$. On CityWalk, the increase is more modest at $12\%$. Unseen performance, however, remains comparable to Seen across datasets, suggesting that the pretrained model does generalize within familiar motion manifolds. This contrast in performance between Counterfactual and Unseen indicates a fundamental limitation in pretrained video world model's ability to extrapolate geometric structure to counterfactual rollouts, a key property expected of grounded world simulators.

\paragraph{O2: Translation and rotation rewards improve spatial alignment with largest gains under counterfactual motion.}
Introducing translation and rotation rewards $r_{\text{trans}}$ and $r_{\text{rot}}$ (\textbf{T+R}) improves spatial alignment across all datasets. On CODa, T+R decreases translation error by $20\%$ in Seen and by $29\%$ in Counterfactual relative to baseline, while rotation error improves by $19\%$ in Seen. On SCAND, the improvements are substantial: translation error decreases by $15\%$ in Seen and by $20\%$ in Counterfactual, with rotation error improving by $4\%$ in Seen. CityWalk shows the strongest counterfactual gains, with Counterfactual translation error dropping by $63\%$ relative to baseline, while Seen improves by $24\%$. These results demonstrate that explicit pose-based feedback enhances motion consistency and stabilizes rollouts $\hat{x}_{1:T}$ across diverse embodiments. Notably, the improvement in Counterfactual highlights that verifiable motion alignment encourages generalization to directionally inverted action sequences $a_{0:T-1}$, an essential feature for embodied reasoning. 

Table~\ref{tab:tr_reliability} further demonstrates that GRPO training systematically improves model reliability: the baseline model shows high variance across both translation and rotation errors, indicating unstable rollouts sensitive to diffusion noise. With \method{} training, both mean errors and variance decrease substantially: at 200 iterations, translation error means reduce by $77\%$ relative to baseline across experiments, with standard deviations reduced by $75\%$. Rotation error also improves, with means reducing by 39\% and standard deviations reduced by $32\%$, achieving reliable and consistent rollouts across all evaluation regimes.

\begin{table*}[h!]
\centering
\begin{threeparttable}
\setlength{\tabcolsep}{4pt}
\renewcommand{\arraystretch}{1.15}

% Column pattern: Method, then T and R for each of Seen, Counterfactual, Unseen
\begin{tabular}{l c c c c c c}
\toprule
& \multicolumn{2}{c}{\textbf{Seen}} & \multicolumn{2}{c}{\textbf{Counterfactual}} & \multicolumn{2}{c}{\textbf{Unseen}} \\
\cmidrule(lr){2-3}\cmidrule(lr){4-5}\cmidrule(lr){6-7}
\textbf{Method}
& \multicolumn{1}{c}{$T\downarrow$}
& \multicolumn{1}{c}{$R\downarrow$}
& \multicolumn{1}{c}{$T\downarrow$}
& \multicolumn{1}{c}{$R\downarrow$}
& \multicolumn{1}{c}{$T\downarrow$}
& \multicolumn{1}{c}{$R\downarrow$} \\
\midrule
Baseline \cite{nvidia2025cosmosworldfoundationmodel} & $73.2 \pm 243.7$ & $2.38 \pm 3.88$ & $75.8 \pm 253.9$ & $2.38 \pm 3.90$ & $71.2 \pm 251.2$ & $2.88 \pm 4.28$ \\
\method{} T+R 100 & $72.0 \pm 283.6$ & $1.95 \pm 3.38$ & $75.9 \pm 311.7$ & $1.85 \pm 3.22$ & $58.4 \pm 231.1$ & $2.57 \pm 4.08$ \\
\method{} T+R 150 & $26.7 \pm 101.5$ & $1.54 \pm 2.84$ & $24.8 \pm 99.1$ & $1.53 \pm 2.80$ & $26.5 \pm 104.3$ & $2.08 \pm 3.33$ \\
\rowcolor{methodcolor!30}
\method{} T+R 200 & $18.4 \pm 68.1$ & $1.40 \pm 2.49$ & $16.8 \pm 63.0$ & $1.36 \pm 2.57$ & $16.3 \pm 56.5$ & $1.97 \pm 3.11$ \\
\bottomrule
\end{tabular}

\vspace{-7pt}
\captionsetup{font=footnotesize}
\caption{Reliability analysis showing error statistics (mean $\pm$ standard deviation) across multiple stochastic rollouts for different GRPO iterations. Baseline exhibits high variance, while GRPO training progressively reduces both mean errors and variance, achieving consistent rollouts. Metrics: T (Translation Error, m), R (Rotation Error, rad).}
\label{tab:tr_reliability}
\end{threeparttable}
\end{table*}

\paragraph{O3: Depth reward enforces local coherence but trades off global alignment.}
Adding the depth reprojection reward $r_{\text{dtr}}$ (\textbf{T+R+DTRI}) enforces local geometric coherence but introduces a trade-off in global rollout alignment. On CODa, compared to T+R, translation error increases by $42\%$ in Seen and by $14\%$ in Counterfactual, while Unseen benefits substantially with a $22\%$ improvement. On SCAND, Seen shows minimal change, Counterfactual degrades by $15\%$, while Unseen improves by $10\%$. CityWalk shows a different pattern where DTRI maintains or slightly improves performance across most regimes. This suggests the depth reward enforces short-horizon consistency and local geometric smoothness, with benefits most apparent in unseen scenarios. 

\paragraph{O4: Full reward set produces most balanced and robust performance.}
Incorporating perceptual feedback through the full reward set (\textbf{T+R+DTRI+V}) produces the most balanced and robust performance across all datasets. On CODa, the full objective achieves translation error reductions of $31\%$ (Seen), $43\%$ (Counterfactual), and $45\%$ (Unseen) relative to baseline, with rotation error improving by $28\%$ (Seen), $8\%$ (Counterfactual), and $11\%$ (Unseen). On SCAND, translation error reduces by $28\%$ (Seen), $30\%$ (Counterfactual), and $5\%$ (Unseen), while CityWalk shows consistent improvements with Counterfactual achieving a $64\%$ reduction. The full objective recovers and extends the rollout accuracy of T+R while preserving local stability from DTRI. Perceptual alignment via video-based evaluators acts as a stabilizer, promoting rollouts that are both physically consistent and visually coherent.
Figure~\ref{fig:qualitative_combined} showcases qualitative comparisons of \method{} rollouts $\hat{x}_{1:T}$ against the baseline. \method{} significantly mitigates the scene drift failures frequently observed when queried with mirrored input action sequences $a_{0:T-1}$, and improves rollout consistency.

\subsection{Training Insights and Stability}

\paragraph{I1: Pretraining bias requires full fine-tuning for meaningful motion adaptation.}
Cosmos-Predict2's pretraining on videos with mostly short clips and slight forward movements biases the model toward scene stability over dynamics. This pretraining bias explains why we observed negligible changes in motion behavior when using parameter-efficient methods like LoRA (except full-size variants), suggesting that pretrained world models may require substantial capacity updates to adapt to navigation-specific action distributions.
\paragraph{I2: Early diffusion timesteps are most critical for GRPO training stability.}
\method{} training is significantly more memory intensive than SFT, as it requires maintaining computation graphs over all diffusion steps to track per-timestep log-probabilities for likelihood ratio computation. Early timesteps are primarily responsible for content denoising while later timesteps refine details with exponentially smaller likelihood values, making the 60\% timestep limitation both memory-efficient and beneficial for training stability by focusing on the most relevant denoising steps.
\paragraph{I3: \method{} requires sufficient reward variance and careful checkpoint selection.}
Most critically, \method{} training stability depends on having sufficient reward variance across stochastic rollouts. \method{} focuses optimization on the reward component with highest variance within each group, which is beneficial for addressing scene drift when some rollouts fail visual odometry while others succeed. However, when checkpoints are overfitted or undertrained, producing uniformly poor or uniformly good rollouts, the normalization step in advantage calculation can amplify subtle differences, leading to degraded training where the model learns arbitrary patterns that add noise to predictions. We mitigate this through KL regularization toward the pretrained model and careful checkpoint selection, ensuring the baseline checkpoint generates distinguishably good and bad rollouts before applying GRPO.

\section{Limitations}\label{sec:limitations}

% We identify multiple limitations. First, our SFT and GRPO ablation experiments were performed over single datasets with limited iterations and batch sizes due to compute constraints; an ideal world model should be jointly trained over multiple datasets over larger iterations. Additionally, we kept simple action-conditioning by our baseline architecture without adopting camera-conditioned pose embeddings or supervised losses; our framework is designed to be agnostic to specific designs of world models, and we leave more sophisticated ablation over multiple baselines as future work. 

RLWG introduces a new paradigm for grounding world models using verifiable rewards, and our study focuses on establishing core principles rather than exhaustively scaling the approach. Our experiments use moderate budgets and limited datasets, but already demonstrate that meaningful grounding can be achieved efficiently. A practical constraint of \method{} is its reliance on sufficient rollout variance for stable relative-policy optimization; understanding the variance dynamics of reward-aligned world model training remains an open direction. Additionally, while RLWG optimizes a multi-objective reward, we do not explore alternative weighting strategies or adaptive weighting schemes, which could further shape the trade-off between global alignment, local consistency, and visual fidelity. Investigating principled multi-reward weighting and larger-scale training holds promise for pushing RLWG toward increasingly stable, generalizable, and physically grounded world models.
\section{Conclusions}\label{sec:conc}

Our work demonstrates that \textbf{RLWG} enables pretrained video world models to be effectively grounded in physical structure through self-supervised post-training, without requiring human labels or external simulators. Instantiated as \method{}, the key insight is that verifiable geometric and perceptual rewards, when optimized via relative policy methods, systematically improve spatial coherence while preserving the visual quality that makes these models powerful priors. The most significant improvements emerge under counterfactual scenarios. \method{} achieves substantial improvements under counterfactual rollouts, with up to $64\%$ reduction in translation error, demonstrating that reward-based alignment addresses a fundamental limitation of current world models: their tendency to prioritize visual plausibility over structural consistency. By explicitly optimizing for both, \textbf{RLWG} enables models that are not only visually coherent but also geometrically consistent, opening new possibilities for reliable planning and control in real-world environments.

\clearpage

\section*{Acknowledgements}
We thank NVIDIA and the authors of Cosmos-Predict2~\cite{nvidia2025cosmosworldfoundationmodel} for publicly releasing their code and checkpoints, upon which we build our post-training pipeline. We thank Nikhil Keetha and collaborators for MapAnything~\cite{keetha2025mapanythinguniversalfeedforwardmetric}, and the authors of VideoAlign~\cite{liu2025improving}, for making their code and checkpoints publicly available, enabling our verifiable self-supervised rewards and evaluation metrics. We also thank Amir Bar and collaborators for releasing Navigation World Models~\cite{bar2025navigationworldmodels} code and checkpoints and for kindly responding to our inquiry email. Finally, we thank Cherie Ho for early-stage discussions and Jay Karhade for discussions on geometric consistency. 

{
    \small
    \bibliographystyle{ieeenat_fullname}
    \bibliography{main}
}

\clearpage
\maketitleappendix

\appendix

\section{Qualitative Results}
\label{sec:appendix:vis}

\begin{figure*}
    \centering
    \includegraphics[width=\textwidth]{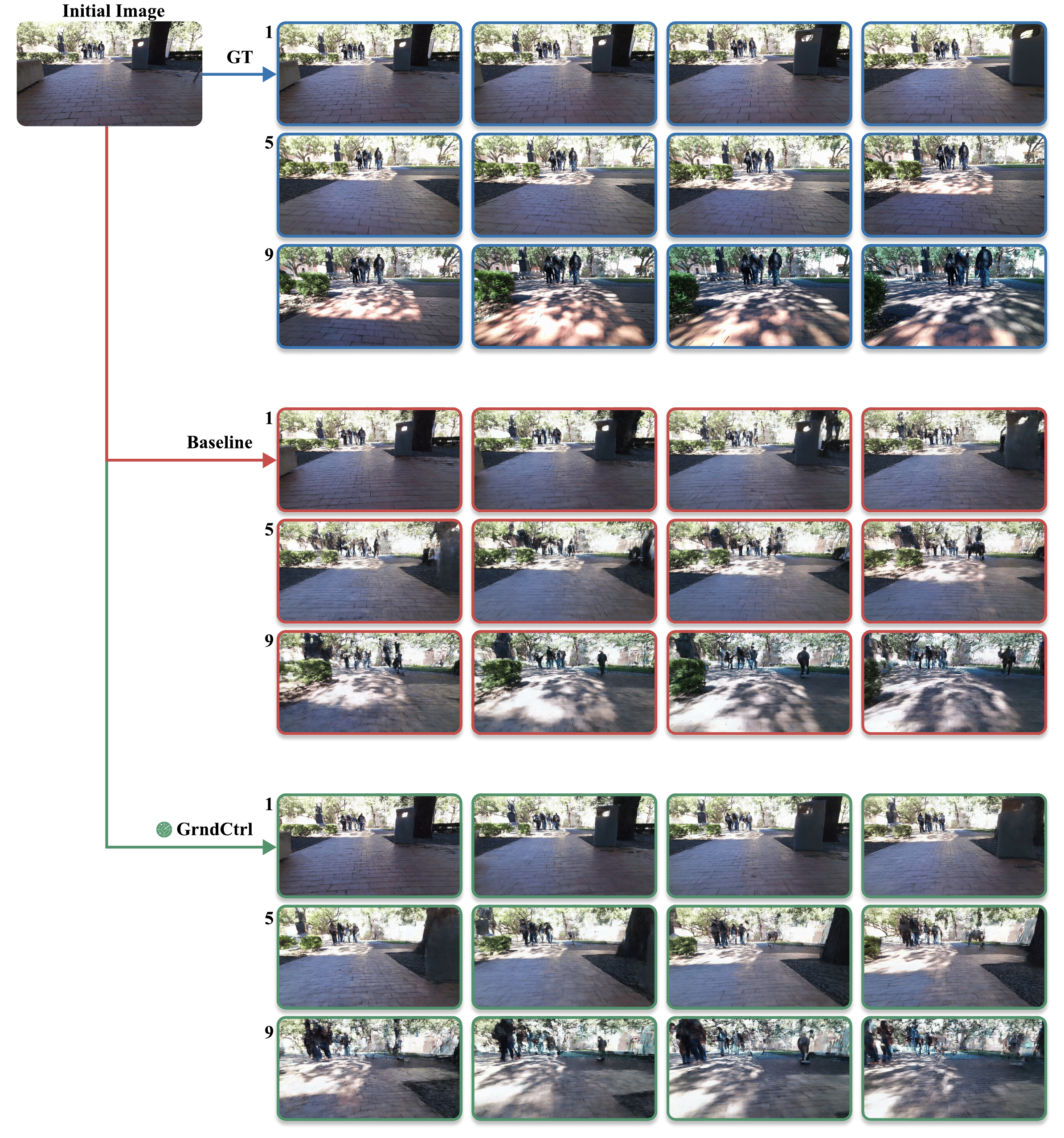}
    \caption{Counterfactual generation example in SCAND. Baseline and \method \ are conditioned on left-right mirrored actions, and the rollouts are expected to invert the movements of the ground truth. Baseline generation still follows similar left-forward movement as GT (trajectory-following failure), while \method \ generation successfully follow inverted right-forward movement.}
    \label{fig:scand11}
\end{figure*}

\begin{figure*}
    \centering
    \includegraphics[width=\textwidth]{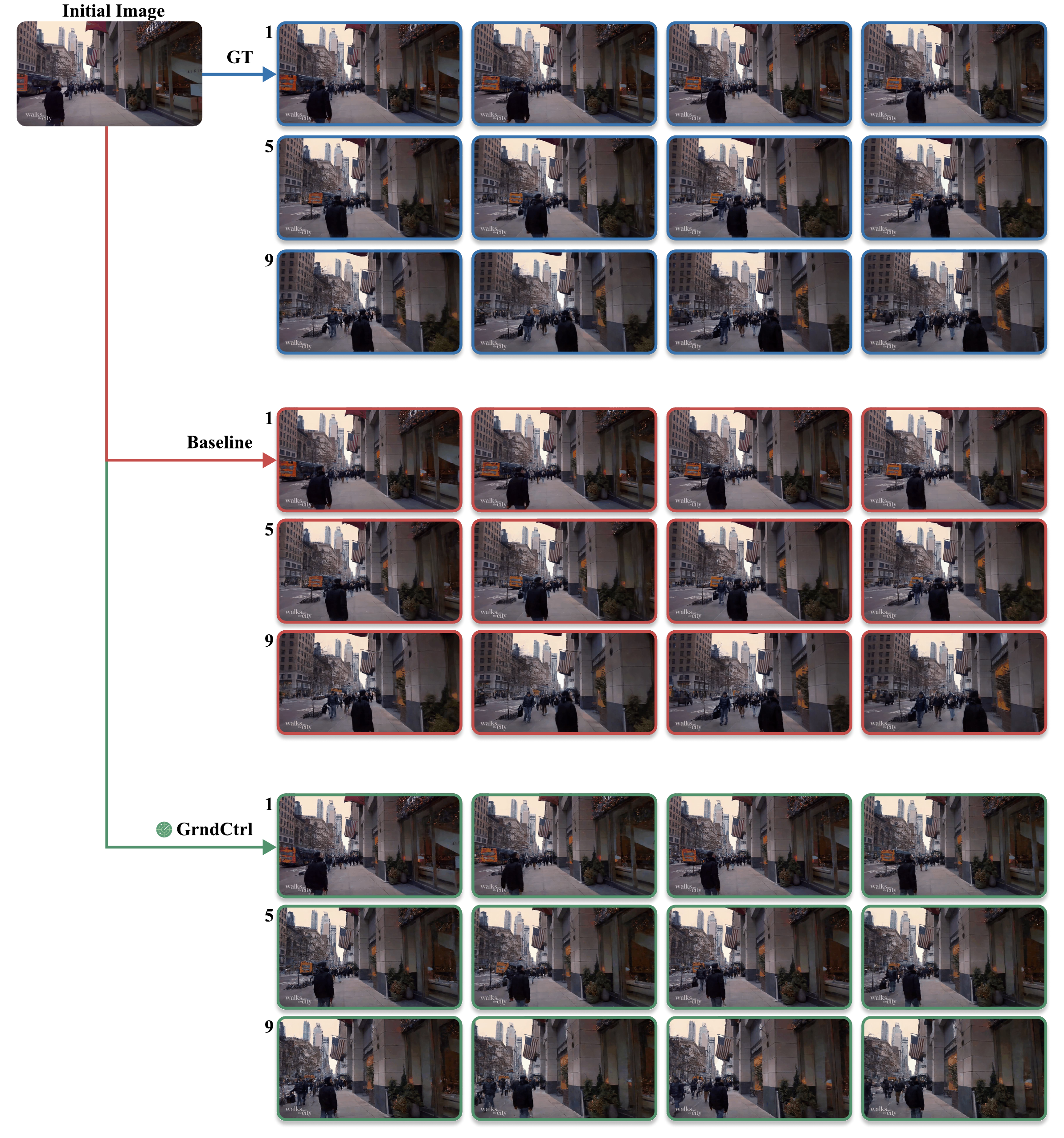}
    \caption{Counterfactual generation example in CityWalk. Baseline and \method \ are conditioned on left-right mirrored actions, and the rollouts are expected to invert the movements of the ground truth. Baseline generation still follows similar left-turn movement as GT (trajectory-following failure), while \method \ generation successfully follow inverted right-turn movement.}
    \label{fig:citywalk71}
\end{figure*}

We include additional counterfactual generation results of \method \ against baseline and ground truths in SCAND~\cite{karnan2022sociallycompliantnavigationdataset} (Figure~\ref{fig:scand11}) and CityWalk~\cite{liu2025citywalker} (Figure~\ref{fig:citywalk71}), in addition to CODa~\cite{zhang2023towards} (Figure~\ref{fig:qualitative_combined}). The counterfactual rollouts of \method \ and baseline are expected to mirror the trajectory shown in ground truths in each scene. We observe \method \ improves trajectory-following and mitigates scene drift in counterfactual rollouts in all three datasets.

% \begin{figure*}
%     \centering
%     \includegraphics[width=\textwidth]{figures/1622.pdf}
%     \caption{Counterfactual generation example of CODa. Baseline and \method \ are conditioned on left-right mirrored actions, and the rollouts are expected to invert the movements of the ground truth.}
%     \label{fig:coda16}
% \end{figure*}

% \begin{figure*}
%     \centering
%     \includegraphics[width=\textwidth]{figures/1310.pdf}
%     \caption{Counterfactual generation example of CODa. Baseline and \method \ are conditioned on left-right mirrored actions, and the rollouts are expected to invert the movements of the ground truth.}
%     \label{fig:coda13}
% \end{figure*}

% \begin{figure*}
%     \centering
%     \includegraphics[width=\textwidth]{figures/Scand62.pdf}
%     \caption{Counterfactual generation example of SCAND. Baseline and \method \ are conditioned on left-right mirrored actions, and the rollouts are expected to invert the movements of the ground truth.}
%     \label{fig:scand62}
% \end{figure*}

% \begin{figure*}
%     \centering
%     \includegraphics[width=\textwidth]{figures/Citywalk230.pdf}
%     \caption{Counterfactual generation example of CityWalk. Baseline and \method \ are conditioned on left-right mirrored actions, and the rollouts are expected to invert the movements of the ground truth.}
%     \label{fig:citywalk230}
% \end{figure*}

\section{Comparisons with NWM \cite{bar2025navigationworldmodels}}
\label{sec:appendix:nwm}

While RLWG is a post-training framework agnostic to the baseline world model, and in \method \  we obtain the baseline with supervised fine-tuning, we acknowledge other potential candidates that perform similar tasks as an experimental baseline. This is exemplified by NWM~\cite{bar2025navigationworldmodels}, which operates a similar world models scenario in navigation tasks. However, significant differences in experimental settings between \method \ and NWM results in infeasible quantitative comparisons:
\begin{itemize}
    \item NWM is jointly-trained on multiple datasets, including RECON~\cite{shah2021rapid} and HuRoN~\cite{hirose2023sacson}, which are collected using fish-eye cameras. This results in frequent twisted artifacts in rollouts conditioned on rectilinear images such as SCAND~\cite{karnan2022sociallycompliantnavigationdataset}. This biases NWM to perform poorly with visual odometry metrics based on a \emph{feed-forward 3D evaluator} such as MapAnything~\cite{keetha2025mapanythinguniversalfeedforwardmetric}. 
    \item \method \ is conditioned on actions defined in 6-DOF space $(x, y, z, \mathrm{roll}, \mathrm{pitch}, \mathrm{yaw})$, whereas NWM's actions are simplified to 3-DOF space $(x, y, \mathrm{yaw})$.
\end{itemize}

\begin{figure*}
    \centering
    \includegraphics[width=0.7\textwidth]{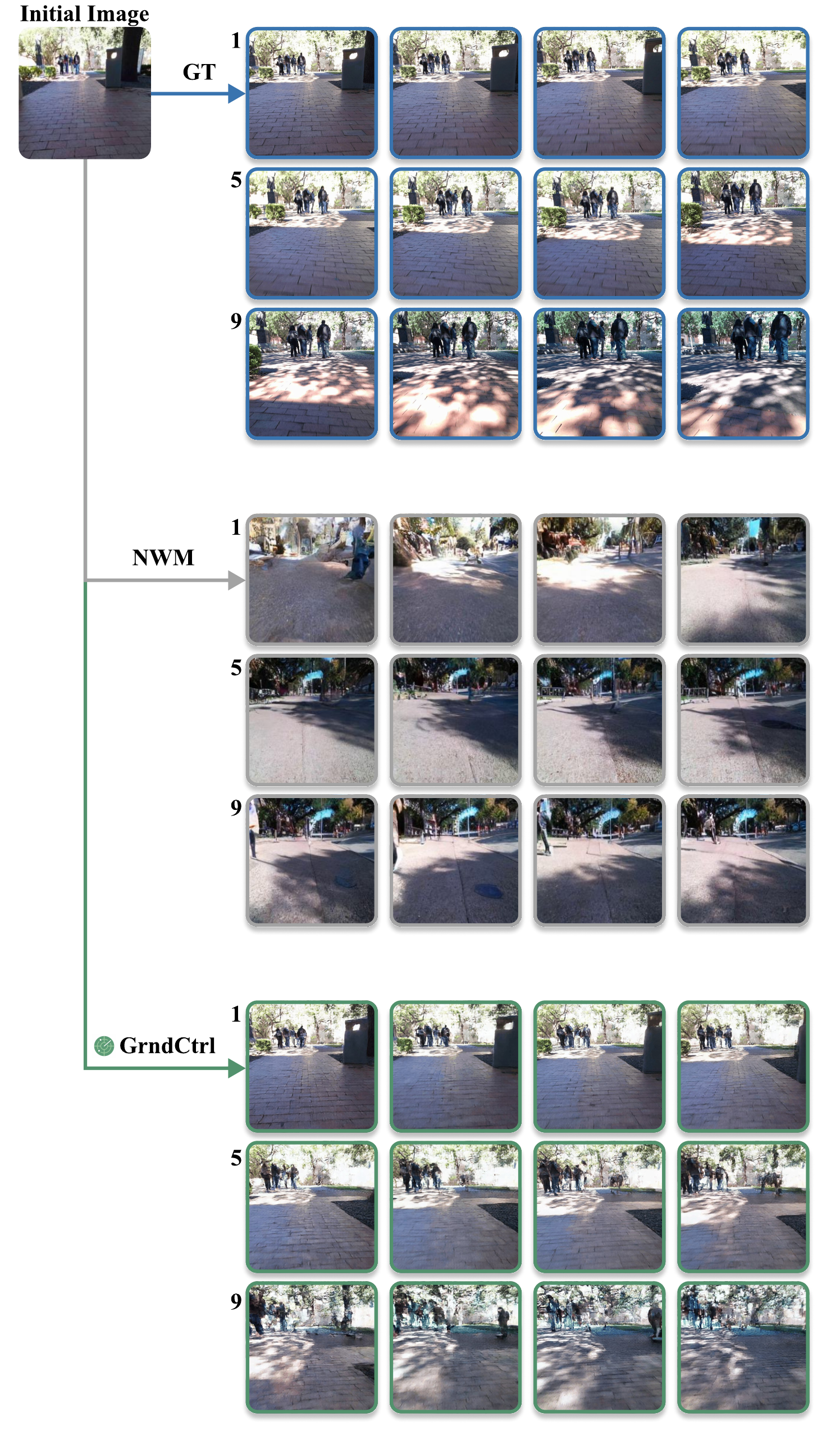}
    \caption{Counterfactual generation example in SCAND. NWM and \method \ are conditioned on left-right mirrored actions, and the rollouts are expected to invert the movements of the ground truth. NWM generation shows clear sign of scene drift in frames 1-3, ending in a more consistent but different scene than the conditional image.}
    \label{fig:nwm}
\end{figure*}

Since \method \ adopts MapAnything as an evaluator to obtain rewards, we are unable to directly use NWM as our baseline pretrained world model $W_\theta$. Nevertheless, we include qualitative samples of NWM counterfactual rollouts in Figure~\ref{fig:nwm}.  We used the same image-action sequences from Figures~\ref{fig:scand11} in SCAND, the dataset shared by both our experiments and NWM. We used 3-DOF actions as conditions, then performed the same inversion for the rest of the trajectory in 3-DOF to generate counterfactual rollouts. We observe similar failures in scene drift and trajectory-following.

\section{Failure Modes}
\label{sec:appendix:failure}

While \method \ demonstrates meaningful grounding in world models, we identify three main failure modes. 

First, the lack of pixel-level supervision in our post-training results in gradual increase in pixel-level noises in our rollouts. While we mitigate this with KL-regularization towards the pretrained model, we still observe a gradual increase in visual noises as our post-training increases overall rewards with by improving trajectory-following or mitigating scene drift in rollouts. When unconstrained in post-training iterations, the added visual noise may eventually exceed the noise tolerance of the \emph{feed-forward 3D evaluator}, resulting in invalid rewards used for training, and the model rollouts gradually collapse to pure noises. 

Second, when a set of rollouts produce similarly good or similarly bad results, our multi-objective normalized group advantage treats the slightly worse good results as negative samples, or the slightly better bad results as positive samples. This results in counterproductive learning in post-training, which causes training instability and gradual collapse to noises. We leave the investigation of rollout variance in world models as future work. 

Third, we occasionally observe reward-hacking behaviors, where rewards increase despite generating noises. Due to the black-box nature of the evaluators and our pipeline stochasticity, we mitigate this by retraining. 
\section{Implementation Details}
\label{sec:appendix:implementation}

We first describe supervised fine-tuning of the baseline world model, followed by GRPO post-training and the evaluator setup. Table~\ref{tab:appendix:hparams} summarizes the key hyperparameters used in both stages.

\subsection{Supervised Fine-Tuning}
\label{sec:appendix:sft}

We fine-tune the Cosmos-Predict2-2B Video2World backbone starting from the released 2B-720p-16fps checkpoint. The visual encoder and decoder of the latent VAE are kept frozen, and we update all parameters in the DiT backbone.

Before settling on this configuration, we experimented with several alternatives: (i) full supervised fine-tuning (all DiT parameters), (ii) LoRA with ranks $16$ and $64$, (iii) a full-size LoRA configuration with $2048 \times 2048$ hidden states, and (iv) training only the action embeddings. Due to Cosmos-Predict2's large-scale pretraining on short clips with mostly mild forward motion, the model is strongly biased toward scene stability rather than rich trajectory dynamics. In practice, only full SFT and the full-size LoRA variant produced meaningful changes in motion behavior; lower-rank LoRA and action-only updates had negligible effect. For simplicity and robustness, we therefore adopt full SFT for all reported experiments.

Supervision is applied in the latent space using an MSE loss with EDM regularization. During SFT, we perform single-step diffusion and backpropagate only through the DiT backbone, which keeps training computationally efficient. We train for $20$k steps with Fully Sharded Data Parallel (FSDP) across eight A100 GPUs. All remaining optimization and diffusion hyperparameters follow the default action-conditioned Cosmos-Predict2 post-training configuration and are listed in Table~\ref{tab:appendix:hparams}.

\subsection{GRPO Post-Training}
\label{sec:appendix:grpo}

GRPO post-training is substantially more GPU-memory intensive than SFT because it requires multiple full video rollouts $\hat{x}_{1:T}$ per conditioning context while maintaining computation graphs over all diffusion steps to track per-timestep log-probabilities. To fit within memory constraints, we train with a batch size of $1$ per GPU on eight A100 GPUs and compensate by sampling $G = 8$ stochastic rollouts for each context $c = (x_0, a_{0:T-1})$, which provides sufficient diversity for group-relative advantage estimation.

A single GRPO update accumulates gradients through the diffusion trajectory to compute per-step likelihood ratios. However, diffusion timesteps do not contribute equally: early steps primarily denoise the main scene content, whereas later steps refine fine-grained details and contribute exponentially smaller likelihood values. Computing likelihood ratios over all steps causes the cumulative log-likelihood to vanish and destabilizes training. In practice, we compute the per-step likelihood ratio only over the first $60\%$ of diffusion timesteps while still running the full sampler, which improves both numerical stability and memory efficiency.

\begin{table}[t]
    \centering
    \small
    \setlength{\tabcolsep}{4pt}
    \begin{threeparttable}
    \begin{tabular*}{\linewidth}{@{\extracolsep{\fill}} c c c @{}}
        \toprule
        Component & Hyperparameter & Value \\
        \midrule
        \multicolumn{3}{c}{\textbf{Optimization}} \\
        \midrule
        SFT/GRPO   & Optimizer                & fused AdamW \\
        SFT/GRPO   & Learning rate            & $1.0 \times 10^{-4}$ \\
        SFT/GRPO   & Weight decay             & $0.1$ \\
        SFT/GRPO   & Betas                    & $(0.9, 0.99)$ \\
        SFT/GRPO   & Epsilon                  & $1.0 \times 10^{-8}$ \\
        SFT/GRPO   & LR schedule              & constant \\
        \midrule
        \multicolumn{3}{c}{\textbf{Diffusion and sampling}} \\
        \midrule
        SFT/GRPO   & Precision                & bfloat16 \\
        SFT        & Classifier-free guidance & $7.0$ \\
        SFT/GRPO   & Diffusion timesteps      & $35$ \\
        SFT        & EDM loss scale           & $10.0$ \\
        SFT        & EDM $\sigma_{\text{cond}}$   & $1.0 \times 10^{-4}$ \\ 
        SFT        & EDM $\sigma_{\text{data}}$   & $1.0$ \\ 
        SFT        & EDM high-$\sigma$ ratio     & $0.0$ \\ 
        GRPO       & Classifier-free guidance & $0.0$ \\
        GRPO       & Timesteps backpropagated & $21$ \\
        \midrule
        \multicolumn{3}{c}{\textbf{Batching and hardware}} \\
        \midrule
        SFT/GRPO   & GPUs                     & $8 \times$ A100 \\
        SFT        & Batch size / GPU         & $8$ \\
        SFT        & Effective batch size     & $64$ \\
        GRPO       & Batch size / GPU         & $1$ \\
        GRPO       & Effective batch size     & $8$ \\
        GRPO       & Rollouts per context $G$ & $8$ \\
        \bottomrule
    \end{tabular*}
    \end{threeparttable}
    \captionsetup{font=footnotesize}
    \caption{Key hyperparameters for supervised fine-tuning (SFT) and GRPO post-training.}
    \label{tab:appendix:hparams}
\end{table}

\subsection{Counterfactual Actions}
\label{sec:appendix:action}

We define the action conditioning of our model as $(x, y, z, \mathrm{roll}, \mathrm{pitch}, \mathrm{yaw})$. In practice, we use the absolute poses with respect to the conditional image's camera frame for each frame we generate. We follow the standard camera coordinate convention with $x$ right, $y$ down, and $z$ forward.

We perform left-right mirroring of the ground truth actions to obtain the counterfactual actions. Mirroring a trajectory with respect to the image center corresponds to reflecting motion across the plane $x = 0$ and flipping the viewing direction around the optical axis. For the translational part, this reflection is $(x, y, z)^\top \;\mapsto\; (-x, y, z)^\top$, which inverts only the lateral component. For small angular magnitudes, the viewing direction can be linearized as $\mathbf{d}(\mathrm{pitch}, \mathrm{yaw}) \approx (\mathrm{yaw}, \mathrm{pitch}, 1)^\top$, so mirroring around the optical axis sends $\mathbf{d} \mapsto (-\mathrm{yaw}, -\mathrm{pitch}, 1)^\top$, corresponding to $(\mathrm{pitch}, \mathrm{yaw}) \mapsto (-\mathrm{pitch}, -\mathrm{yaw})$. Combining these, a 6-DOF action $(x, y, z, \mathrm{roll}, \mathrm{pitch}, \mathrm{yaw})$ has the counterfactual action $(-x, y, z, \mathrm{roll}, -\mathrm{pitch}, -\mathrm{yaw})$,
which is the mirror of the original motion with respect to the conditioning image’s center axis.

This method of obtaining counterfactual actions ensure realistic movement of the embodiment as those in the datasets. When conditioning using arbitrary actions, the model rollouts from the pretrained model are more likely to have poorer quality. 

\subsection{Evaluators and Rewards}
\label{sec:appendix:evaluators}

MapAnything~\cite{keetha2025mapanythinguniversalfeedforwardmetric} serves as our frozen 3D evaluator $E$, providing relative pose estimates $(\Delta R_t, \Delta \mathbf{t}_t)$ and per-frame depth maps $D_t$ for each rollout $\hat{x}_{1:T}$. VideoAlign~\cite{liu2025improving} serves as our frozen video evaluator $V$, providing sequence-level visual quality, motion quality, and text alignment scores. We use only visual and motion quality scores (with equal weighting $\alpha=\beta=0.5$) for the video quality reward $r_{\text{v}}$, ignoring text alignment since our rollouts are not text-conditioned. When GPU memory is constrained, model parameters of the evaluators are temporarily offloaded to CPU, and re-loaded only when used.

\end{document}